\documentclass[letterpaper, 10 pt, conference]{ieeeconf}  % Comment this line out if you need a4paper

\IEEEoverridecommandlockouts                              % This command is only needed if 
\usepackage{graphicx} % for pdf, bitmapped graphics files
\usepackage[usenames,dvipsnames,table]{xcolor}
\usepackage{cite} % 引用のパッケージを追加
\usepackage{framed}

\usepackage{graphicx} % for pdf, bitmapped graphics files
\usepackage{subcaption}
\usepackage[usenames,dvipsnames,table]{xcolor}
\usepackage{amsmath}
\usepackage{amssymb}

\makeatletter
    \let\NAT@parse\undefined
\makeatother
\usepackage[square,numbers,sort&compress]{natbib}

\usepackage{hyperref}
\hypersetup{
    colorlinks=true,
    bookmarks=true,
    linkcolor=MidnightBlue,
    filecolor=magenta,   
    urlcolor=MidnightBlue,
    citecolor=MidnightBlue,
}

\usepackage[capitalise, nameinlink]{cleveref}

\title{\LARGE \bf
SPARK: Graph-Based Online Semantic Integration System for Robot Task Planning
}

\author{Mimo Shirasaka$^{1}$, Yuya Ikeda$^{1}$, Tatsuya Matsushima$^{1}$, Yutaka Matsuo$^{1}$ and Yusuke Iwasawa$^{1}$% Albert Author$^{1}$ and Bernard D. Researcher$^{2}$% <-this % stops a space
\thanks{$^{1}$The University of Tokyo. Email: m-shirasaka@g.ecc.u-tokyo.ac.jp
}
}
\begin{document}

\maketitle
\thispagestyle{empty}
\pagestyle{empty}

%%%%%%%%%%%%%%%%%%%%%%%%%%%%%%%%%%%%%%%%%%%%%%%%%%%%%%%%%%%%%%%%%%%%%%%%%%%%%%%%
\begin{abstract}

The ability to update information acquired through various means online during task execution is crucial for a general-purpose service robot. This information includes geometric and semantic data. While SLAM handles geometric updates on 2D maps or 3D point clouds, online updates of semantic information remain unexplored. We attribute the challenge to the online scene graph representation, for its utility and scalability. Building on previous works regarding offline scene graph representations, we study online graph representations of semantic information in this work. We introduce SPARK: Spatial Perception and Robot Knowledge Integration. This framework extracts semantic information from environment-embedded cues and updates the scene graph accordingly, which is then used for subsequent task planning. We demonstrate that graph representations of spatial relationships enhance the robot system's ability to perform tasks in dynamic environments and adapt to unconventional spatial cues, like gestures.

\end{abstract}

%%%%%%%%%%%%%%%%%%%%%%%%%%%%%%%%%%%%%%%%%%%%%%%%%%%%%%%%%%%%%%%%%%%%%%%%%%%%%%%%
\section{INTRODUCTION}

% \vspace{-0.25\baselineskip}
For a robot to generate actions that reflect the time-shifting semantics of a dynamic environment, it must consider object features, the uncertainty of geometric information over time, and the varying reliability of information acquired differently. The challenge of managing and updating this crucial information for decision-making can be attributed to online scene graph updates, due to their utility and scalability. During task execution, sensors can acquire various types of environmental information, including the locations of objects and humans and environment-intrinsic tendencies. This information falls into two categories: geometric and semantic. Online updates of geometric information correspond to map updates used for motion planning. However, online updates of semantic information remain unexplored. While geometric information aids in task planning, it may not always adapt effectively. For instance, if a child places an apple on a laundry machine, the robot might struggle to find it without understanding the semantic relationship. Knowing that the apple is on the laundry machine, an uncommon location, is essential for effective task execution.

In previous studies, 3D-LLM~\citep{hong20233d} introduces a framework that uses 3D geometric data for task planning, while ConceptGraphs~\citep{gu2023conceptgraphs} presents an open-vocabulary, graph-structured 3D scene representation method for robot perception and planning. However, in both works, semantic information sourced from other than geometric representation is not reflected in the scene representation update. Self-Recovery Prompting~\citep{shirasaka2023self} leverages both geometric and semantic data for task planning, but their updates are not simultaneous.

\begin{figure}[t]
    \centering
    \includegraphics[width=\linewidth]{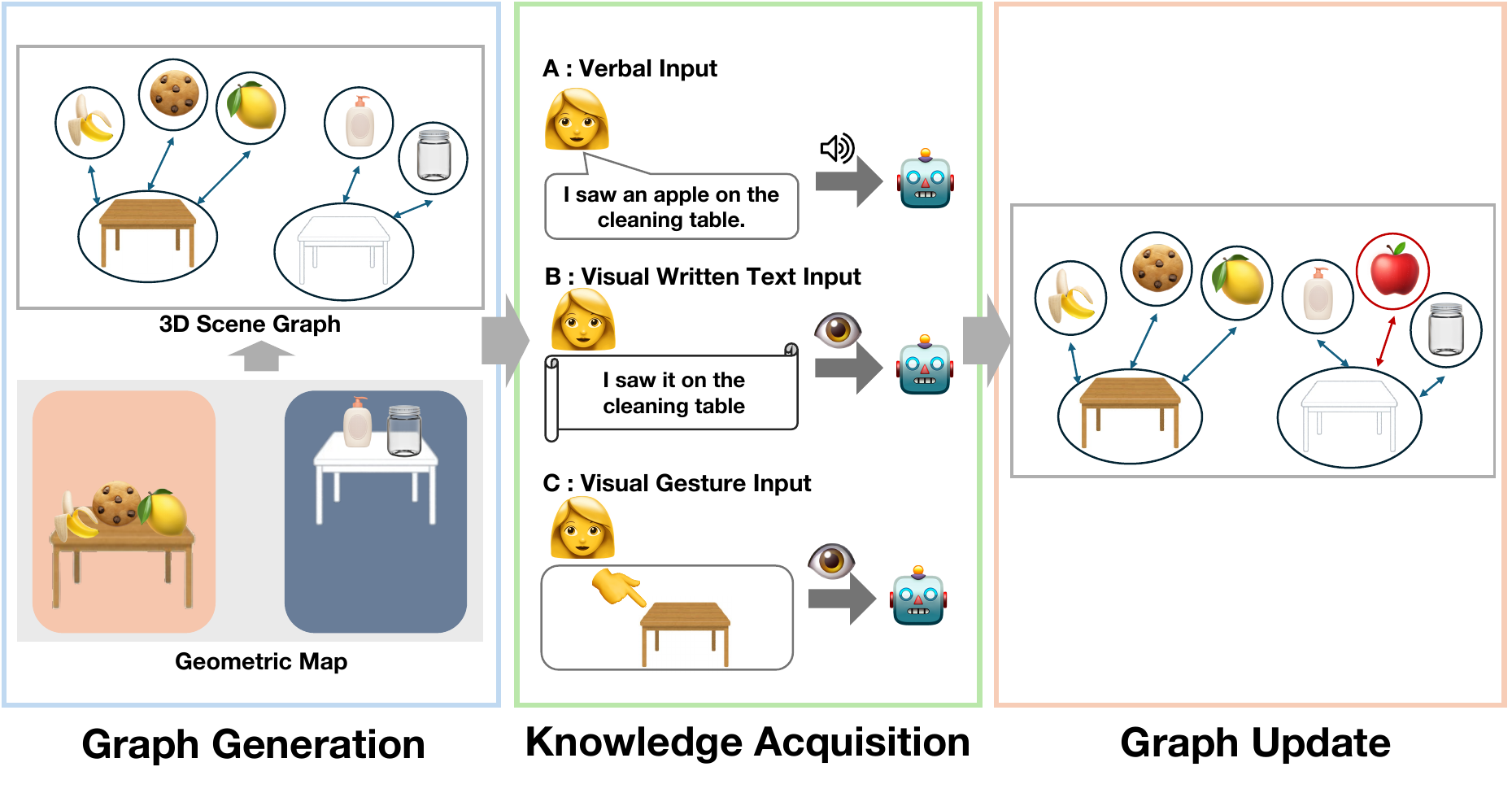}
    \caption{Overview of the SPARK system. The system extracts semantic information from geometric data and environment-embedded cues, updating the scene graph online for further task planning.}
    \label{fig:system_overview}
\end{figure}

To integrate environment-embedded cues into semantic information and reflect this in online task planning, we introduce SPARK: Spatial Perception and Robot Knowledge Integration for general-purpose service robots. This framework, illustrated in~\autoref{fig:system_overview},  incorporates text-based information and geometric data from the environment into online planning. Environment-embedded cues, such as signs, human speech, and human gestures, provide semantic information about positional relationships. SPARK extracts semantic information from these cues and updates the scene graph, which is then used for further task planning.

The text-based information includes human speech, signs, signals, and gestures. Previously, the inability to update this information stemmed from a lack of semantic data acquisition and the failure to incorporate updates. Upon detecting a person during task execution, SPARK seeks information relevant to the ongoing task from the individual. When semantic information is obtained, the system updates its graph of semantic representation and, depending on the content of the new information, remakes the plan based on the updated graph.

% This template provides authors with most of the formatting specifications needed for preparing electronic versions of their papers. All standard paper components have been specified for three reasons: (1) ease of use when formatting individual papers, (2) automatic compliance to electronic requirements that facilitate the concurrent or later production of electronic products, and (3) conformity of style throughout a conference proceedings. Margins, column widths, line spacing, and type styles are built-in; examples of the type styles are provided throughout this document and are identified in italic type, within parentheses, following the example. Some components, such as multi-leveled equations, graphics, and tables are not prescribed, although the various table text styles are provided. The formatter will need to create these components, incorporating the applicable criteria that follow.

\section{SYSTEM OVERVIEW}
\subsection{General Purpose Service Robot System}
We target the application of our system for the realization of a general-purpose service robot, where tasks are performed in a household environment and the robot is expected to perform tasks given verbally by the operator. The following four foundation models enhance the system's ability to be generalized and adaptive with prompting: Whisper~\citep{whisper} for speech recognition, GPT-4~\citep{gpt4} for task planning, Detic~\citep{detic} for object detection and segmentation, and CLIP~\citep{clip} for object classification. By integrating these models into our system, we establish an entirely promptable baseline for this work.

\begin{figure}[t]
    \centering
    \includegraphics[width=\linewidth]{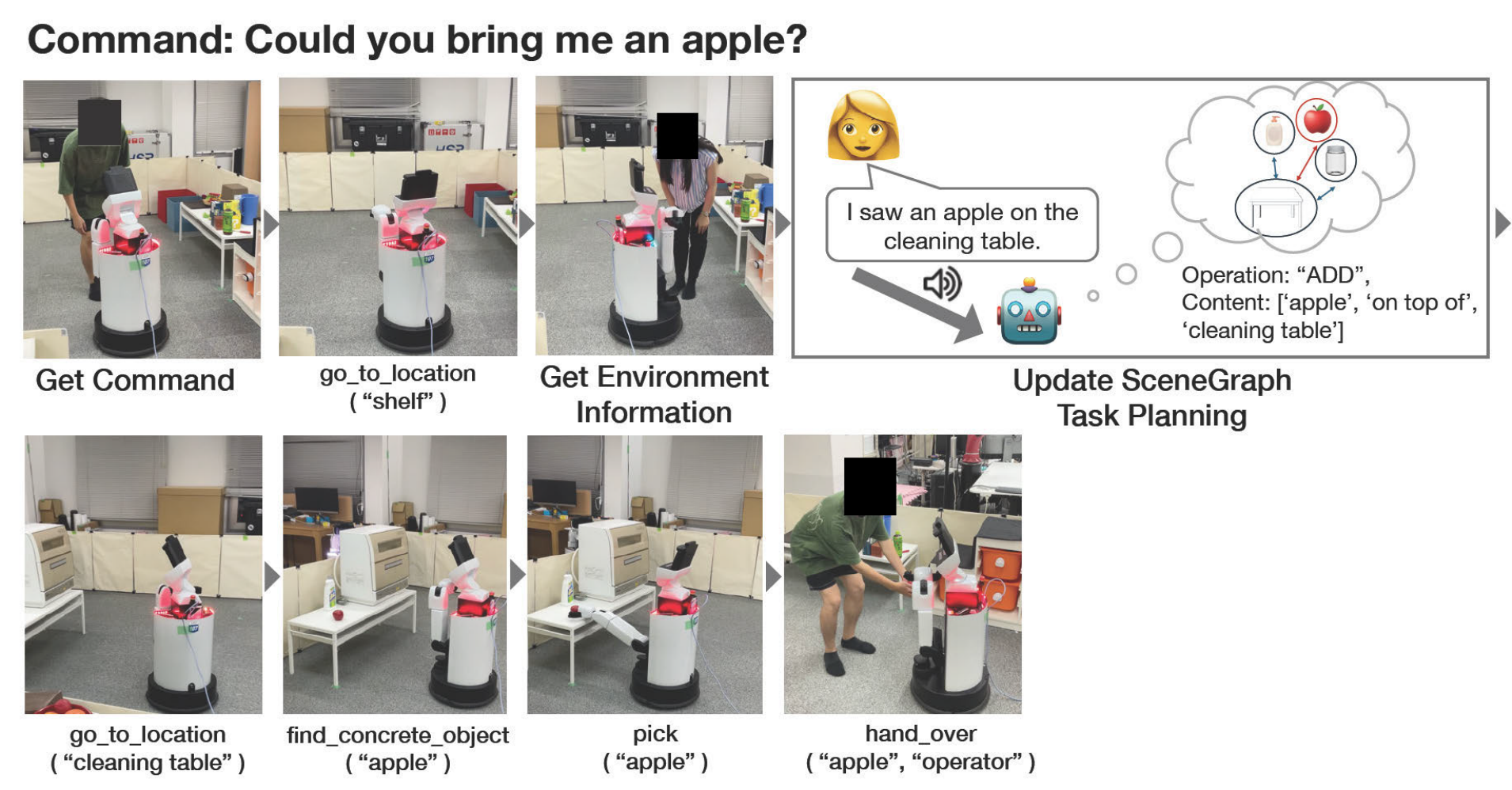}
    \caption{Example of verbal input. The robot plans to go to the \textit{``shelf''} for an \textit{``apple''} but changes to the \textit{``cleaning table''} based on verbal information.}
    \label{fig:1_verbal}
\end{figure} 

\subsection{SPARK System}

To address the challenge of integrating environment-embedded cues into semantic information for online task planning, we introduce SPARK, a framework depicted in~\autoref{fig:system_overview}. We validate SPARK with experiments using examples of transcribable semantic information, such as \textit{verbal input}, \textit{written text input}, \textit{gesture input}, and \textit{geometric information} from SLAM.

\subsubsection{Geometric to Semantic Conversion}

By adapting the ConceptGraphs framework \citep{gu2023conceptgraphs} for online updates, the system converts geometric information into semantic relationships during task execution. This allows the robot's scene graph to be updated automatically and leveraged to adjust the execution plan in mid-performance.

\subsubsection{Other Embedded Cues to Semantic Conversion}

When receiving verbal input, we query the LLM to decide on the next steps: updating the map and/or updating the plan. If the newly acquired semantic information is relevant to the ongoing task, the system updates the scene graph and the execution plan accordingly. When receiving visual input, we query the LLM to extract semantic information. Based on the LLM output, our framework updates the scene graph and adjusts task planning as necessary.

\section{EXPERIMENTS AND RESULTS}
Experiments were conducted to evaluate the system’s ability to update the task execution plan upon receiving additional information about the task environment using our proposed approach. We used the Human Support Robot (HSR) developed by Toyota Motor Corporation~\citep{hsr} in a real-world simulated household environment, consisting of a living room and a cleaning room.
\subsection{Verbal Input}
The experiment in \autoref{fig:1_verbal} demonstrates the robot obtaining semantic information from a person verbally. The robot was commanded to bring an \textit{``apple''} without being given its specific location. In the task environment, the apple was located on a \textit{``cleaning table''}, an unlikely place according to the LLM task planner in the original system, potentially causing the robot to fail. However, the developed system overcame this by integrating the \textit{``apple''}’s location into the task planning and successfully completing the command.

\begin{figure}[t]
    \centering
    \includegraphics[width=\linewidth]{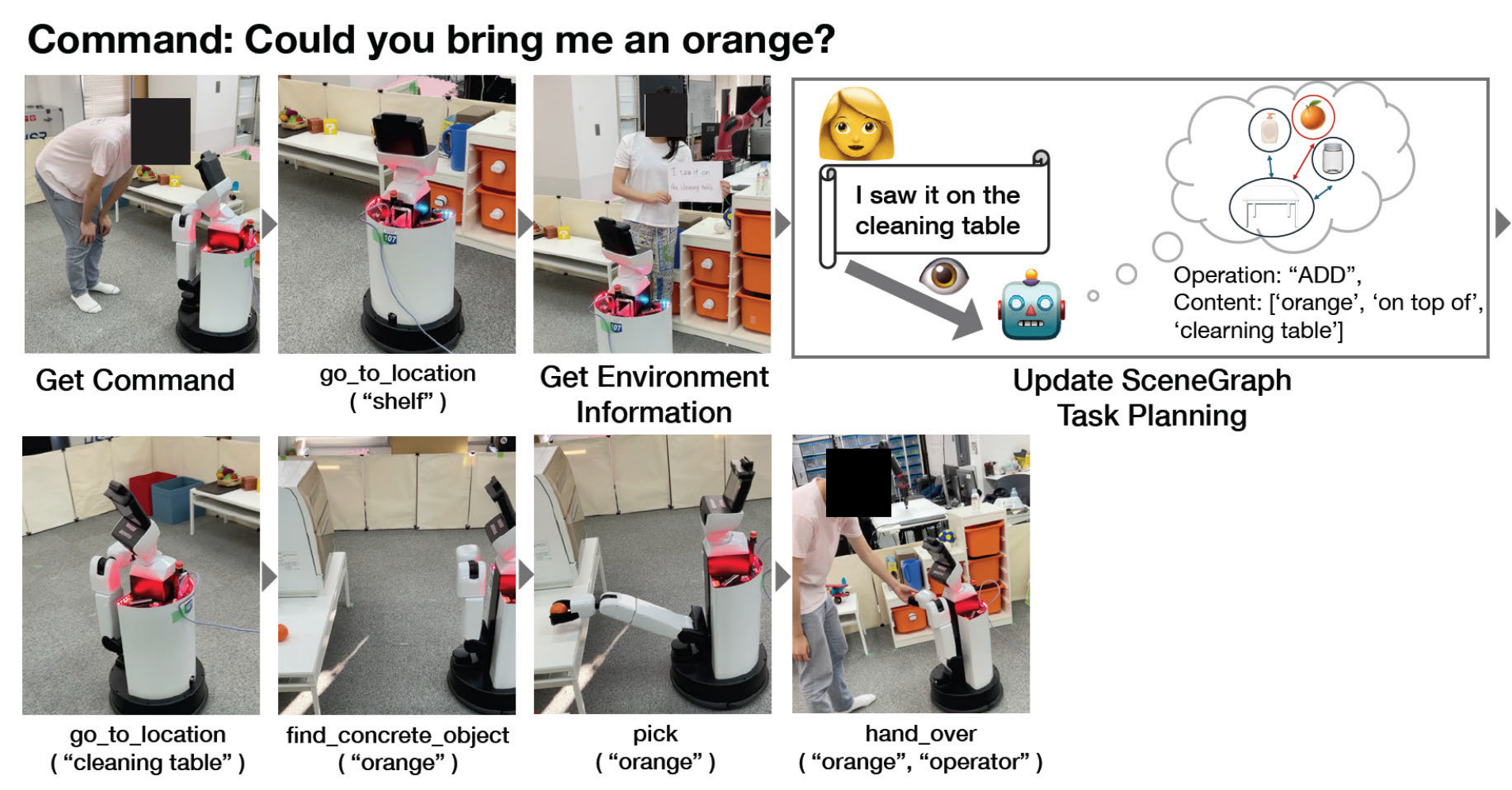}
    \caption{Example of written text input. The robot plans to go to the \textit{``shelf''} for an \textit{``orange''} but changes to the \textit{``cleaning table''} based on visually detected written information.}
    \label{fig:2_writtentext}
\end{figure} 
\begin{figure}[t]
    \centering
    \includegraphics[width=\linewidth]{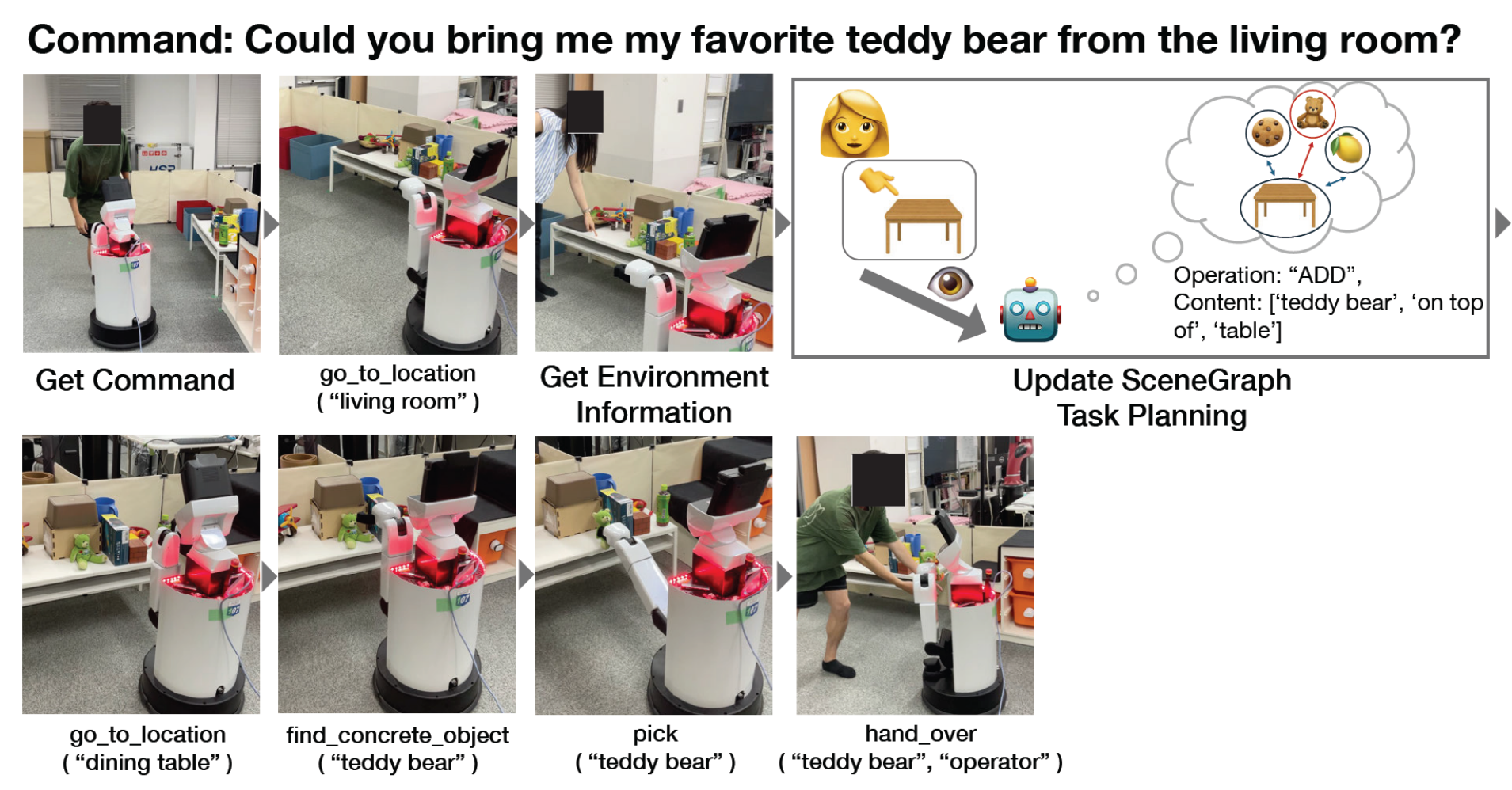}
    \caption{Example of gesture input. The robot plans to go to the \textit{``living room''} for a \textit{``teddy bear''} but changes to the \textit{``dining table''} based on visually detected gesture information.}
    \label{fig:3_gesture}
\end{figure} 

\subsection{Written Text Input}
\autoref{fig:2_writtentext} shows the robot obtaining task-related semantic information from visual context in written text format. The commanded object was an \textit{``orange''}, and other settings were nearly identical, except for the method of additional information acquisition. The robot was shown written text indicating the \textit{``orange''} is on the \textit{``cleaning table''}. By updating the scene graph and reflecting it in task planning, the robot completed the command.

\subsection{Gesture Input}
The final experiment in \autoref{fig:3_gesture} shows the robot obtaining task-related information from a gesture (a human pointing at the \textit{``dining table''}). The robot was commanded to bring a \textit{``teddy bear''} from the \textit{``living room''}. Although the teddy bear was in the \textit{``living room''}, the abstract indication could have caused the robot to take longer to find it. By interacting with a detected person, the robot acquired more specific information about the \textit{``teddy bear''}'s location from gestures, updated the scene graph and adjusted its actions accordingly.

\section{CONCLUSION}
We proposed SPARK: Spatial Perception and Robot Knowledge Integration, a framework that extracts semantic information from environment cues and updates the scene graph online for task planning. We demonstrated that graph representations of spatial relationships help the robot perform tasks in dynamic environments and adapt to unconventional spatial cues like gestures.

\bibliographystyle{IEEEtranN}
\bibliography{main}

% Generated by IEEEtranN.bst, version: 1.14 (2015/08/26)
\begin{thebibliography}{8}
\providecommand{\natexlab}[1]{#1}
\providecommand{\url}[1]{#1}
\csname url@samestyle\endcsname
\providecommand{\newblock}{\relax}
\providecommand{\bibinfo}[2]{#2}
\providecommand{\BIBentrySTDinterwordspacing}{\spaceskip=0pt\relax}
\providecommand{\BIBentryALTinterwordstretchfactor}{4}
\providecommand{\BIBentryALTinterwordspacing}{\spaceskip=\fontdimen2\font plus
\BIBentryALTinterwordstretchfactor\fontdimen3\font minus \fontdimen4\font\relax}
\providecommand{\BIBforeignlanguage}[2]{{%
\expandafter\ifx\csname l@#1\endcsname\relax
\typeout{** WARNING: IEEEtranN.bst: No hyphenation pattern has been}%
\typeout{** loaded for the language `#1'. Using the pattern for}%
\typeout{** the default language instead.}%
\else
\language=\csname l@#1\endcsname
\fi
#2}}
\providecommand{\BIBdecl}{\relax}
\BIBdecl

\bibitem[Hong et~al.(2023)Hong, Zhen, Chen, Zheng, Du, Chen, and Gan]{hong20233d}
Y.~Hong, H.~Zhen, P.~Chen, S.~Zheng, Y.~Du, Z.~Chen, and C.~Gan, ``3d-llm: Injecting the 3d world into large language models,'' \emph{Advances in Neural Information Processing Systems}, vol.~36, pp. 20\,482--20\,494, 2023.

\bibitem[Gu et~al.(2023)Gu, Kuwajerwala, Morin, Jatavallabhula, Sen, Agarwal, Rivera, Paul, Ellis, Chellappa, et~al.]{gu2023conceptgraphs}
Q.~Gu, A.~Kuwajerwala, S.~Morin, K.~M. Jatavallabhula, B.~Sen, A.~Agarwal, C.~Rivera, W.~Paul, K.~Ellis, R.~Chellappa \emph{et~al.}, ``Conceptgraphs: Open-vocabulary 3d scene graphs for perception and planning,'' \emph{arXiv preprint arXiv:2309.16650}, 2023.

\bibitem[Shirasaka et~al.(2023)Shirasaka, Matsushima, Tsunashima, Ikeda, Horo, Ikoma, Tsuji, Wada, Omija, Komukai, et~al.]{shirasaka2023self}
M.~Shirasaka, T.~Matsushima, S.~Tsunashima, Y.~Ikeda, A.~Horo, S.~Ikoma, C.~Tsuji, H.~Wada, T.~Omija, D.~Komukai \emph{et~al.}, ``Self-recovery prompting: Promptable general purpose service robot system with foundation models and self-recovery,'' \emph{arXiv preprint arXiv:2309.14425}, 2023.

\bibitem[Radford et~al.(2023)Radford, Kim, Xu, Brockman, McLeavey, and Sutskever]{whisper}
A.~Radford, J.~W. Kim, T.~Xu, G.~Brockman, C.~McLeavey, and I.~Sutskever, ``Robust speech recognition via large-scale weak supervision,'' in \emph{International Conference on Machine Learning}.\hskip 1em plus 0.5em minus 0.4em\relax PMLR, 2023, pp. 28\,492--28\,518.

\bibitem[{OpenAI}(2023)]{gpt4}
{OpenAI}, ``{GPT-4 Technical Report},'' \emph{arXiv e-prints}, p. arXiv:2303.08774, Mar. 2023.

\bibitem[Zhou et~al.(2022)Zhou, Girdhar, Joulin, Kr{\"a}henb{\"u}hl, and Misra]{detic}
X.~Zhou, R.~Girdhar, A.~Joulin, P.~Kr{\"a}henb{\"u}hl, and I.~Misra, ``Detecting twenty-thousand classes using image-level supervision,'' in \emph{European Conference on Computer Vision}.\hskip 1em plus 0.5em minus 0.4em\relax Springer, 2022, pp. 350--368.

\bibitem[Radford et~al.(2021)Radford, Kim, Hallacy, Ramesh, Goh, Agarwal, Sastry, Askell, Mishkin, Clark, et~al.]{clip}
A.~Radford, J.~W. Kim, C.~Hallacy, A.~Ramesh, G.~Goh, S.~Agarwal, G.~Sastry, A.~Askell, P.~Mishkin, J.~Clark \emph{et~al.}, ``Learning transferable visual models from natural language supervision,'' in \emph{International conference on machine learning}.\hskip 1em plus 0.5em minus 0.4em\relax PMLR, 2021, pp. 8748--8763.

\bibitem[Yamamoto et~al.(2019)Yamamoto, Terada, Ochiai, Saito, Asahara, and Murase]{hsr}
T.~Yamamoto, K.~Terada, A.~Ochiai, F.~Saito, Y.~Asahara, and K.~Murase, ``Development of human support robot as the research platform of a domestic mobile manipulator,'' \emph{ROBOMECH journal}, vol.~6, no.~1, pp. 1--15, 2019.

\end{thebibliography}
\clearpage

\end{document}